%% file: root.tex
\documentclass[letterpaper, 10 pt, conference]{ieeeconf}  

\IEEEoverridecommandlockouts                              

\usepackage{graphics} 
\usepackage{epsfig} 
\usepackage{hyperref}
\usepackage{mathptmx} 
\usepackage{times} 
\usepackage{amsmath} 
\usepackage{amssymb}  
\usepackage{todonotes}
\usepackage{booktabs}
\input{meta/usercmds}
\usepackage{multicol}
\usepackage{multirow}
\graphicspath{{image/}}
\usepackage{todonotes}
\usepackage{svg}
\let\labelindent\relax
\usepackage{enumitem}
\newcommand\Tstrut{\rule{0pt}{2.6ex}}         
\newcommand\Bstrut{\rule[-0.9ex]{0pt}{0pt}} 
\title{\LARGE \bf
L-DYNO: Framework to Learn Consistent Visual Features Using Robot's Motion.
}

\author{Kartikeya Singh, Charuvaran Adhivarahan and Karthik Dantu
\thanks{All authors are from the Department of Computer Science and Engineering.
        University at Buffalo, Buffalo, NY (emails:{ ksingh35, charuvah, kdantu} @buffalo.edu)
        }%
\thanks{Corresponding Author: ksingh35@buffalo.edu}
\thanks{\textcolor{blue}{\href{https://github.com/kartikeya13/L-DYNO}{Project Page}}}
}

\begin{document}

\maketitle
\thispagestyle{empty}
\pagestyle{empty}

\begin{abstract}
Historically, feature-based approaches have been used extensively for camera-based robot perception tasks such as localization, mapping, tracking, and others. Several of these approaches also combine other sensors (inertial sensing, for example) to perform combined state estimation. Our work rethinks this approach; we present a representation learning mechanism that identifies visual features that best correspond to robot motion as estimated by an external signal. Specifically, we utilize the robot's transformations through an external signal (inertial sensing, for example) and give attention to image space that is most consistent with the external signal. We use a pairwise consistency metric as a representation to keep the visual features consistent through a sequence with the robot's relative pose transformations. This approach enables us to incorporate information from the robot's perspective instead of solely relying on the image attributes. We evaluate our approach on real-world datasets such as KITTI \& EuRoC and compare the refined features with existing feature descriptors. We also evaluate our method using our real robot experiment. We notice an average of \textbf{49\%} reduction in the image search space without compromising the trajectory estimation accuracy. Our method reduces the execution time of visual odometry by \textbf{4.3\%} and also reduces reprojection errors. We demonstrate the need to select only the \textit{most important features} and show the competitiveness using various feature detection baselines.

\end{abstract}

\section{INTRODUCTION}

 Crucial robotics applications including search and rescue, agricultural robotics, industrial automation, and self-driving cars heavily rely on the robot's ability to localize itself in complex environments. Visual sensors such as monocular, stereo, and depth cameras are popular sensing modalities for perception. Prior works have utilized an understanding of sensor physics to detect features in the sensor readings for use in robot perception. An example is the use of image features for visual odometry and mapping. However, a challenge with this methodology is identifying errors from the sensor or environmental factors that affect the perception. This work develops a representation learning framework that selects sensor features (image features, for example) that best correspond with robot motion as sensed through another sensor. 

Feature-based methods like RANSAC \cite{raguram2008comparative}, TEASER++ \cite{yang2020teaser} are sensitive to image features. Unique features produce consistent matches and transformations with such techniques, which lower rotation and translation errors. However, non-unique features are always present in real-world datasets due to various reasons. For example, outdoor datasets have similar-looking objects like trees, sky, cars, and buildings. Ignoring these non-unique features would improve the transformation estimation process in algorithms like bundle adjustment. In this work, we propose a learning-based algorithm that reduces the image features to a subset of the most consistent features, likely most suitable for downstream perception tasks. 

Metrics from feature-based methods cannot be used for learning due to two reasons: (i) using signals from the same sensor would not reduce errors. Ideally, we would like to use a signal from a different sensor and (ii) these feature-matching functions are non-continuous. Threshold-based feature matching methods have zero gradients almost everywhere and are undefined for some inputs. Fortunately, a plethora of signals in robotics that are independent, continuous, and are good indicators of consistency. In this work, we use IMU-based transformations as our consistency signal to train our network. Rather than use an end-to-end learning like~\cite{jiang2021cotr,mao20223dg} to find feature matches directly, we modify the input images highlighting areas for the best feature-matching so that we can rely on robust and well-tested methods in computer vision and robotics to do the feature-matching in the reduced space.

Using features that are indicative of the estimates is just as important as feature density in reducing estimation errors. For instance, consistent features extracted from a distinct portion of the scene, even in a lesser number, can assist in a better-estimated trajectory. Apart from highlighting the most relevant regions in the scene for feature matching, our approach reduces the dimensionality of the input search space. To handle the drift in the IMU-based transformations, we use a window-based approach \cite{wei2021unsupervised} to obtain the IMU-based pose transformations, which help in supervising the consistency-based loss shown in figure~\ref{fig:overview}. We evaluate our method on real-world autonomous driving datasets such as \kitty~\cite{geiger2013vision} and \euroc~\cite{burri2016euroc} and compare them with the popular feature detector baselines. We also evaluate our method using a custom dataset generated from a real robot car based upon F1tenth \cite{o2020f1tenth} setup. 

The main contributions of this work are as follows:
\begin{itemize}[leftmargin=1em,topsep=0.1em,itemsep=0.05em]
  \item We introduce an attention-based deep learning architecture to determine which regions of a scene are important for consistent feature detection.
  \item We take advantage of the robot's motion with an IMU-based loss function in our learning module. We generate a representation between inertial sensing and consistent image features by utilizing IMU consistency in extracting \textit{most important} image space.
  \item We benchmark our approach by performing evaluations on various real-world datasets \kitty~\cite{geiger2013vision}, \euroc~\cite{burri2016euroc}, and our dataset recorded from a robot. We use different feature detector baselines (both classical and learning-based) and show the reduced number of outliers being produced using our method. 
  \item We compare our method with the baselines and show an improvement in Average Trajectory Error per baseline shown in table~\ref{table3} and a reduction in Average Execution Time by \textbf{4.3\%} by reducing the image space up to \textbf{49\%}.
  \item  We further evaluate our image space by calculating reprojection errors of all the baseline feature detectors. We notice a reduction in the average reprojection error which leverages accurate homography estimation for pose recovery. 
\end{itemize}

\section{Related Work}\label{sec:related_work}
\subsection{Classical and Learning-Based Feature Detectors}
Traditional feature detection and matching methods are still being used to achieve accurate relative pose estimation through tracking visual features. Most VO methods depend on the visual features from images using feature extraction techniques like \cite{derpanis2004harris,harris1988combined,carron1994color} and perform tracking using geometrically aligned methods such as LK-Tracking \cite{baker2004lucas}, brute force matching with KNN \cite{garcia2010k} and FLANN-based matching \cite{vijayan2019flann}. These techniques are highly sensitive to noise, which may result in an inaccurate estimation of a trajectory when influenced by a significant number of outliers. Once features are detected, several geometrical techniques such as relative pose estimation or epipolar geometry-based triangulation of 3D points can be used to estimate an accurate pose trajectory. However, these geometric-based methods lack robustness over large datasets and with time the odometry accumulates drift \cite{younes2017keyframe}. Furthermore, various SLAM systems such as ORB-SLAM \cite{mur2015orb}, Lsd-SLAM \cite{engel2014lsd}, Svo \cite{forster2014svo} and Dso \cite{engel2017direct} perform reasonably well in both detecting and computing features followed by estimating the trajectory in an end-to-end fashion.

With the advancements in the domain of deep learning for SLAM, researchers have been attempting to replace classical methods of detecting and tracking features with learning-based methods. In terms of feature tracking, learning-based VO methods perform well \cite{mohanty2016deepvo,wang2021tartanvo,gao2022unsupervised,vijayanarasimhan2017sfm,li2018undeepvo} but many of them rely only on image features and do not consider information from the robot's dynamics, which adversely affects transformation estimates if the tracked features are inconsistent. In this work, we make use of the robot's inertial measurements, i.e. IMU (Inertial Measurement Unit) in the loss function to supervise our training mechanism to track consistent features and improve the generalizability in learning methods \cite{yin2018geonet,wang2019unos}. Many VO frameworks use hand-crafted feature detectors \cite{chiu2013fast,mur2015orb} and learning-based feature descriptors \cite{detone2018superpoint,sun2018pwc,sarlin2020superglue}. However, our method considers using information from the robot's dynamics and uses it as a supervision signal to train our learning pipeline. Most of the feature descriptors incorporate outliers, which hinders an accurate estimation of trajectory. More advanced learning techniques involve attention-based neural networks for feature detection and matching. In contrast, LoFTR \cite{sun2021loftr} COTR \cite{xie2021cotr} E2EMVM \cite{roessle2022end2end} and SUPERGLUE \cite{sarlin2020superglue} present Graphical Neural Networks which perform well in estimating correct feature correspondences with a wide baseline. The receptive fields used in these methods help in determining fine textures in a global context, which sets them apart from standalone CNN-based learning methods.
\subsection{Outlier Rejection Techniques}
Thresholding-based techniques that restrain the flow of outliers in a VO pipeline prevent the large influence of outliers in the system, but the step function used in these techniques hinders the computation of gradients in a learning framework. RANSAC \cite{raguram2008comparative} and TEASER++ \cite{yang2020teaser} are the two examples of strategies used to deal with outliers. RANSAC works by randomly sampling a subset of features detected and fitting a model to this subset. The model is further evaluated by measuring how well it fits the remaining features. This technique may show a non-deterministic property of selecting different data points in every iteration. TEASER++ is a recently introduced method that rejects outliers by incorporating additional techniques such as global optimization and local refinement. Our method relies on the same objective of ignoring outliers but considering only specific attentive image regions. Our attention heads output the attentive blocks that determine the image space required to achieve both feature detection and tracking accuracy.

\section{Method}\label{sec:approach}
\begin{figure}[t]
    \centering
    \includegraphics[width=\columnwidth]{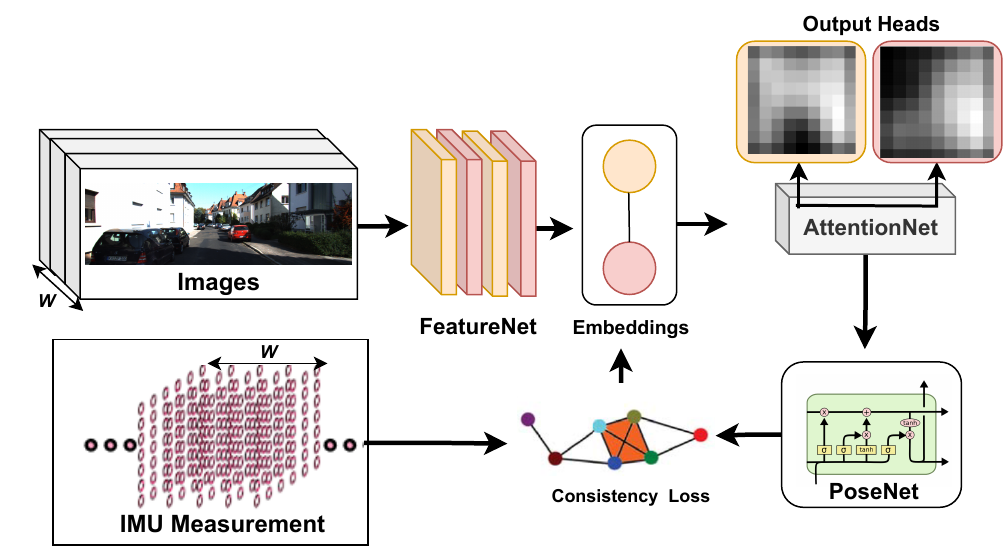}
    \caption{Overview of the Pipeline: FeatureNet takes in the raw image frame pairs and provides feature maps. Feature maps are passed to an Attention Network that assigns weights to certain regions of the image. These learned features go through an LSTM-based PoseNet which provides 6-dof robot pose trained with a consistency-based loss function formulated with IMU transformations.}
    \label{fig:overview}

\end{figure}

The objective of this work is to reduce the number of outlier features that are inconsistent with a motion sensor (IMU). To achieve this, we aim to map the consistent visual features with the IMU-based inertial consistency. For this, we train a deep neural network that reduces the region of interest for feature detection with signals from robot motion in our loss function $L$=$[p^i_w, q^i_w]$. We introduce a differentiable approach to create a feature selection pipeline that takes supervised signals from an IMU. To handle drift accumulated over time using IMU, we follow a window-based approach, which takes small bursts of measurements and creates relative transformation proxies which are inherited consistently throughout the trajectory formation. The overview of our method is depicted in figure~\ref{fig:overview}. In this section, we describe the various components of our method. In section~\ref{sec:evaluation} we show results from our evaluations of our method compared with other feature-based visual odometry methods.

\begin{figure}[b]
    \includegraphics[width=8.5cm, height=3.5cm]{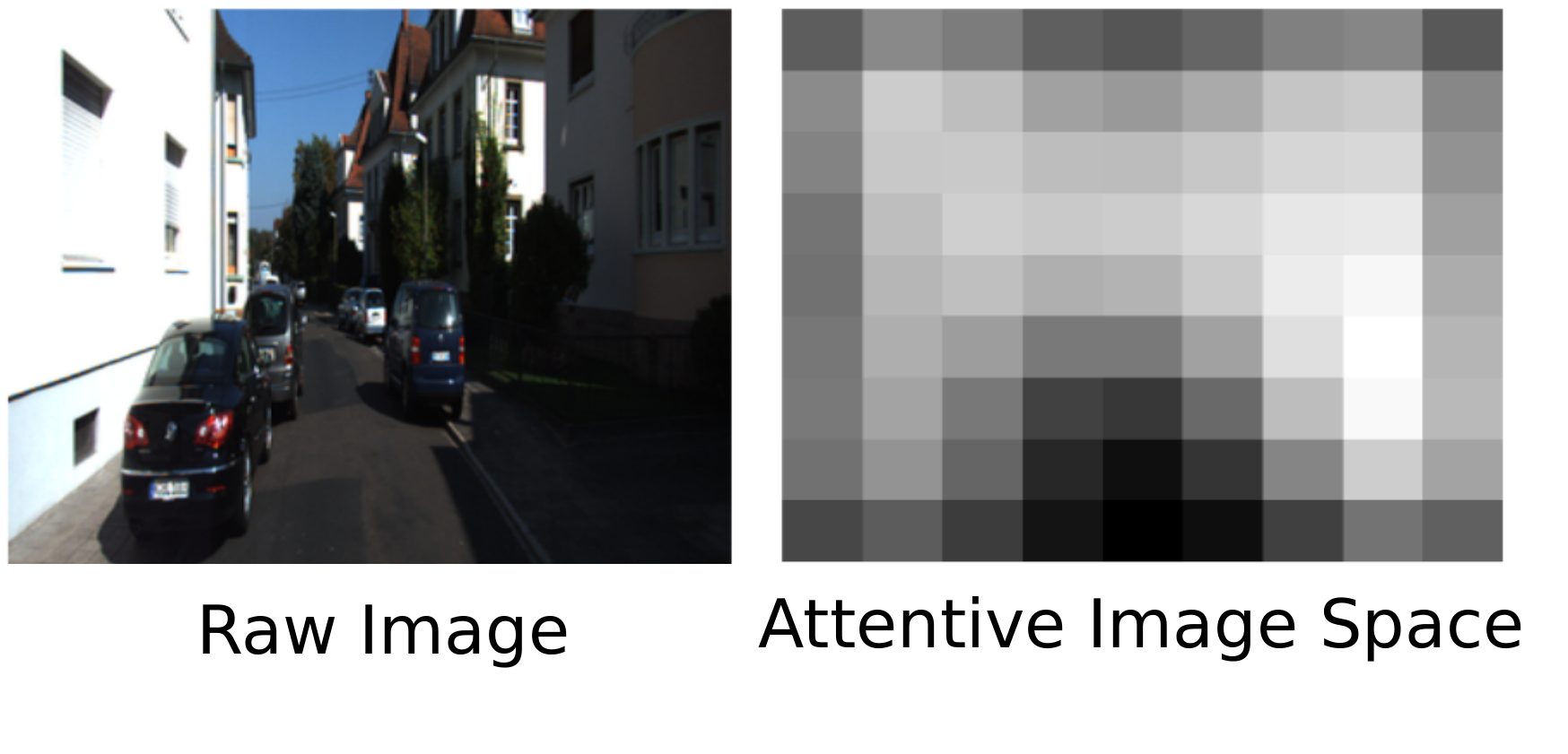}
    \caption{Feature to Descriptor: We obtain attentive features from our AttentionNet output heads, which are further overlayed by the baseline feature detectors. This overlaying of baseline feature detectors on our attentive image space refines the outliers.}
    \label{fig:featuretodescriptor}
\end{figure}

Our training mechanism includes three main modules. We name them as: \textit{FeatureNet}, \textit{PoseNet}, and \textit{AttentionNet}. Figure~\ref{fig:overview} depicts the complete pipeline. Since IMU sensors have noise in their gyroscope and accelerometers, we use a sliding window-based approach that includes a short sequence of image pairs ($t$ to $t+1$ $\forall\, t \in W$  ) and IMU bursts within the same timestamp-based window $(W)$ similar to VIO (Visual and Inertial Odometry) methods \cite{wei2021unsupervised} and this helps reduce the effect of noise from IMU given by:
\begin{equation}\label{first_eq}
       a_m = C_q^{-1}(a_t - g) + b_a - n_a
 \end{equation}
\begin{equation}\label{first_eq}
       w_m = w_t + b_w - n_w
\end{equation}
Where: $b_a$ and $n_a$ are the bias and noise accumulated from the accelerometer due to external factors, $g$ is the gravity vector on the accelerometer and $Cq$ is the 3x3 rotational matrix representing IMU orientation with respect to a solid coordinate system. A similar representation can be seen for a gyroscope, where $w_t$ is the true acceleration void of any external disturbance and $w_m$ is the measured acceleration from the sensor. We exploit these sensors to create a consistent pose transformation and use them to train the following subnetworks:

\textbf{FeatureNet:}
The FeatureNet network includes a multi-layered CNN block which takes the full-size input image pairs and provides two feature maps of size $256 \times M \times N$. In figure~\ref{fig:featuretodescriptor}, we can observe the extracted attentive blocks. We use feature detector points from the baselines to overlay on top of only these attentive blocks, instead of the whole image. We show that this reduced search space extracted using our method reduces translation, rotation, and pose errors in section~\ref{sec:evaluation}. These feature maps correspond to the pairwise embedding vectors per image and store the information about the features extracted. We do not use any pre-trained network as a backbone in any of our training modules. 

\textbf{Attention Network:}
The output features from FeatureNet are passed through a self-attention-based network, which takes shared weights from the embeddings and selects only attentive weights that improve the training stream. These weights are saved and further passed for training in the next batch. As an intermediate step, our AttentionNet provides trained attention heads, which are treated as the image search space and considered as the most important feature space.

\textbf{PoseNet:}
Finally, an LSTM-based pose network takes the attentive features and weights to estimate a stream of 6-dof poses. This whole training mechanism iterates through the attention maps from the AttentionNet and reduces the consistency-based loss.

\textbf{Consistency-based Loss:} \label{loss}
For loss minimization, we make use of the robot's transformation poses over the specified window size. These transformations are from IMU. The estimated IMU can be concluded from the following equations:
\begin{align} \label{eq3-7}
    p_w^i &= v_w^i,\\
    v_w^i &= C_{q_{w}^{i}}(a_m-b_a+n_a)-g,\\
    q_w^i &= \frac{1}{2}\Omega(w_m-b_w+n_w)q_{w}^{i},\\
    b_a &= n_{b_a},\\
    b_w &= n_{b_w}
\end{align}
Equations (3)-(7) are the kinematics model of an IMU transformation. Consider the position of IMU in the World frame $p^i_w$ , the velocity of IMU in the World frame is $v^i_w$ with $dt = 1s$, and the parameter representing the rotation from World coordinate to IMU stationary coordinate is $q^i_w$. We have bias generated due to external factors $b_a$ and $b_w$ in the accelerometer and gyroscope.
Further, $C_{q^i_w}$ represents the rotation matrix that shows the transformation of a vector from the relative frame to the world frame, $\mu$ is the quaternion product matrix and $g$ is the vector due to gravity. The transformed pose of the robot is represented by $[p^i_w, q^i_w]$. The integration of these differential equations helps us estimate the pose from IMU measurements. The PCM-based consistency metric calculates the relative poses $[p^i_w, q^i_w]$ over a window$(W)$ and constructs proxies of the same. Further, we use MSE (Mean Squared Error) to reduce the spatial gap between the 6-dof obtained from our PoseNet module and the 6-dof obtained from $[p^i_w, q^i_w]$.

\section{Evaluation}\label{sec:evaluation}

\begin{figure*}[t]
    \centering
    \includegraphics[width=17.7cm, height=4.5cm]{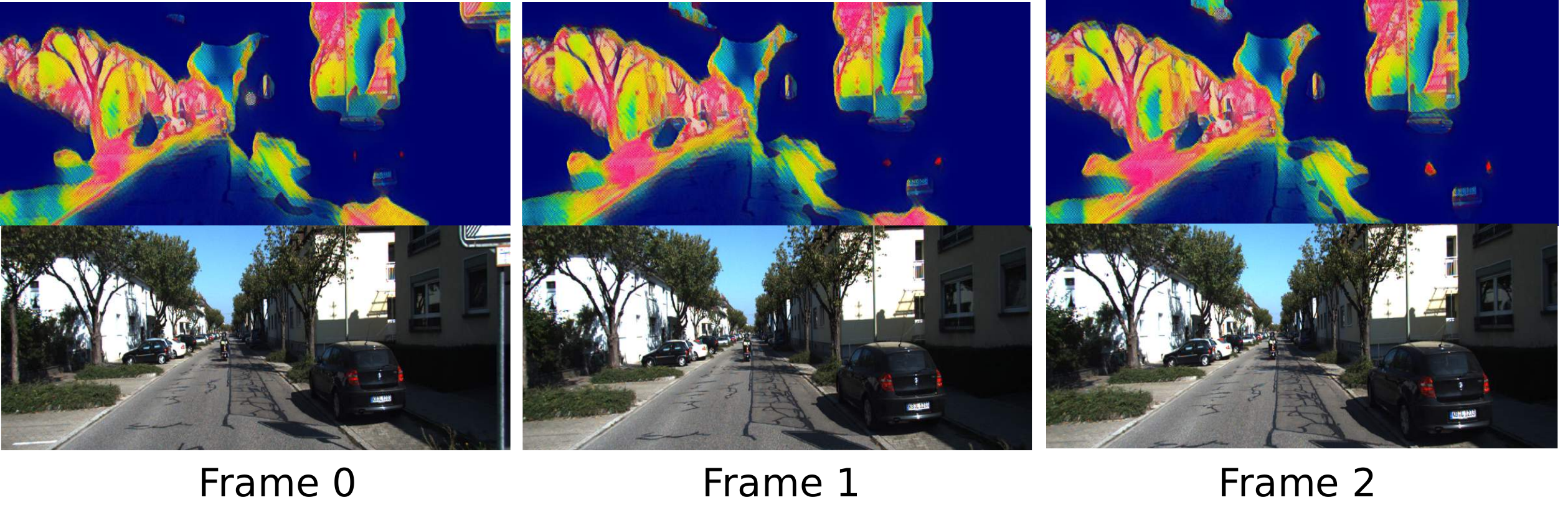}
    \caption{Attentions(above) obtained from the input sequences (S0 from KITTI dataset). We can notice the consistency maintained throughout the images, which results in an accurate pose estimation explained in section.\ref{sec:evaluation}.}
    \label{fig:attentions}
\end{figure*}

\begin{figure}
    \centering
    \includegraphics[width=\columnwidth]{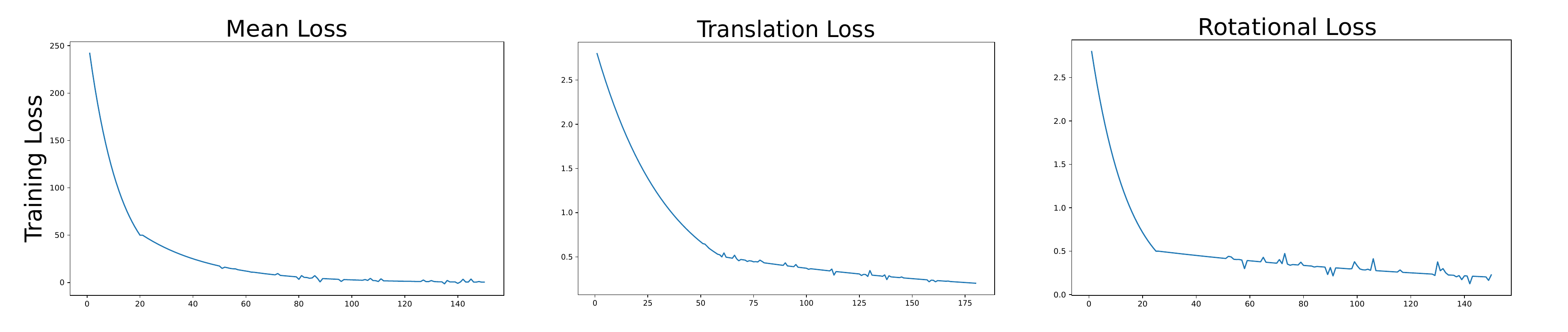}
    \caption{Training R,t loss from the complete pipeline.}
    \label{fig:loss}
        \vspace*{-10pt}
\end{figure}

\textbf{Network Training:} For the experiments, we use KITTI and EuRoC sequences with their raw IMU as a loss function. We train the pipeline end-to-end on the RTX A100 graphics card. We use a standard MSE loss to balance out the weights with our defined loss in section~\ref{sec:approach}. Figure~\ref{fig:loss} represents the loss convergence over time with respect to the reduced error between the PoseNet module and defined IMU-based transformation parameters. We show the associative convergence of loss in table~\ref{table1} which shows the reduced mean error between our IMU-based transformations and resultant 6DOF poses from our PoseNet module. We train our models on full-size input images for both KITTI (376x1241) and EuRoc(480x752). We incorporate full-size images to exploit complete image details in our AttentionNet resulting in feature map generation. These feature maps in figure~\ref{fig:attentions} demonstrate the selective attention regions over a sequential set of input images. In this section, we evaluate the performance of our method with attentive regions with VO-based pose errors, translation and rotation errors, distribution of inliers/outliers, reprojection errors, and execution time. We also demonstrate results using a real-world experiment presented in Section~\ref{sec:f1tenth}. 

\begin{table}[b]
\begin{center}

  \begin{tabular}{|l|l|l|l|l|l|l|}
    \hline
    \multirow{2}{*}{Dataset} &
      \multicolumn{2}{c}{T0} &
      \multicolumn{2}{c}{T1} &
      \multicolumn{2}{c|}{T2} \\
    & t-error & R-error & t-error & R-error & t-error & R-error \\
    \hline

    KITTI \cite{geiger2013vision} \Tstrut\Bstrut & 1.7m & 2.2m & 1.8m & 1.1m & 1.6m & 2.3m \\
    \hline
    EuRoc \cite{burri2016euroc} \Tstrut\Bstrut & 1.6m & 2.6m & 2.3m & 1.9m & 1.6m & 2.2m \\
    \hline
  \end{tabular}\\
\end{center}
\Tstrut\Bstrut
\vspace*{-1em}
\caption{Intermediate rotation and translation errors from the PoseNet module with respect to the IMU transformations provided for loss supervision.}
\label{table1}
\vspace*{-2em}
\end{table}
%
From section~\ref{sec:approach}, we obtain attentive regions shown in figure~\ref{fig:attentions} which associates the feature consistency over a sequence with the masked image space expressed as the region with \textit{Important Features}. As a qualitative analysis, we can notice that the attentive region is consistent over the input sequence shown in figure~\ref{fig:attentions}. The heat maps highlight the area that represents the most attentive region, and we can observe that these are on areas that are intuitively good indicators of relative transformation.
\subsection{Benchmarking between the Inliers and Outliers}
\begin{figure*}
    \centering
    \includegraphics[width=1\textwidth]{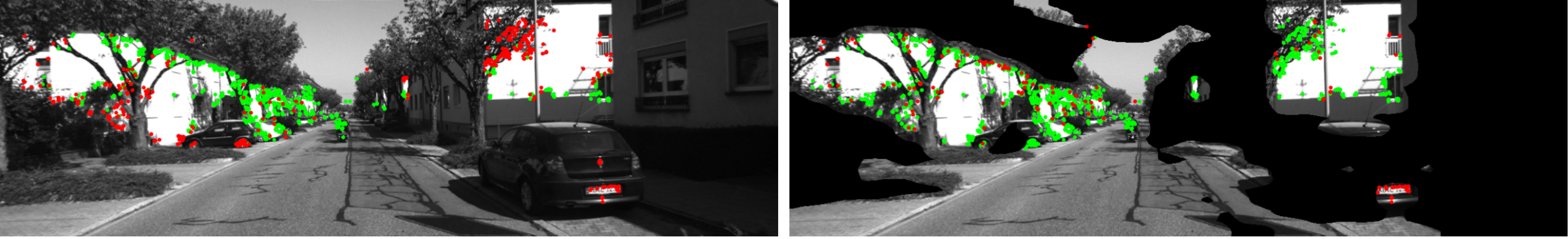}
    \caption {ORB features on the original space(Left) vs ORB features on our reduced space(Right). Our method removes the outliers by reducing the search space area resulting in a more accurate estimation of homography without compromising the transformation accuracy shown in table~\ref{table3}.}
    \label{fig:inlierpic}
\end{figure*}

We show the data association accuracy between the correct features obtained using our method, by comparing them against keypoints generated using current state-of-the-art methods such as ORB \cite{mur2015orb}, SIFT \cite{alhwarin2010vf}, BRISK \cite{leutenegger2011brisk}, AKAZE \cite{sharma2020image}, KAZE \cite{alcantarilla2012kaze} and SuperPoint \cite{detone2018superpoint}. We further compare the distribution of inliers and outliers over the original image space vs our reduced search space using RANSAC. We overlay all the baseline features detected over both the image spaces and obtain inliers vs outliers distribution using RANSAC. Table~\ref{table2} shows the mean and standard deviation of inliers and outliers while using a standalone baseline over the original image space vs using baseline feature detectors over our \textit{Attentive Image Space} obtained with our method. We obtain a cumulative decrease in the number of outliers in our method as compared to most of the baselines. In figure~\ref{fig:inlierpic}, we qualitatively examine the difference in outliers from both image spaces.
\begin{table*}[h]
\begin{center}
\begin{tabular}{|l|l|l|l|l|l|l|l|l|l|}
\hline
\multirow{2}{*}{Detectors} & \multicolumn{4}{c}{KITTI \cite{geiger2013vision}} & \multicolumn{4}{c}{EuRoc \cite{burri2016euroc}} \\
& $\mu$ In & $\sigma$ In & $\mu$ Out & $\sigma$ Out& $\mu$ In & $\sigma$ In & $\mu$ Out & $\sigma$ Out \\
\hline
ORB \cite{mur2015orb} \Tstrut\Bstrut & 882 & 248 & 274 & 98 & 394 & 35 & 214 & 30 \\
\textbf{Ours}   & 587 & 320 & 101 & 92 & 388 & 31 & 109 & 44  \\
\hline
SIFT \cite{alhwarin2010vf} \Tstrut\Bstrut & 998 & 257 & 335 & 81 & 411 & 44 & 302 & 58 \\
\textbf{Ours}   & 1102 & 208 & 114 & 66 & 388 & 43 & 218 & 53  \\
\hline
FAST \cite{muja2012fast} \Tstrut\Bstrut & 912 & 129 & 473 & 137 & 481 & 76 & 498 & 53 \\
\textbf{Ours}    & 822 & 108 & 205 & 136& 460 & 82 &  402& 54  \\
\hline
KAZE \cite{alcantarilla2012kaze} \Tstrut\Bstrut & 675 & 346 & 435 & 84& 397 & 58 & 372 & 61 \\
\textbf{Ours}    & 412 & 237 & 319 & 86& 402 & 73 & 389 & 52  \\
\hline
AKAZE \cite{sharma2020image} \Tstrut\Bstrut & 641 & 241 & 390 & 91 & 367 & 75 & 377 & 56\\
\textbf{Ours}    & 330 & 173 & 252 & 77& 403 & 65 & 395 & 69  \\
\hline
BRISK \cite{leutenegger2011brisk} \Tstrut\Bstrut & 403 & 138 & 390 & 66 & 392 & 54 & 405 & 63\\
\textbf{Ours}    & 241 & 139 & 237 & 73 & 388 & 66 & 363 & 57  \\
\hline
SuperPoint \cite{detone2018superpoint} \Tstrut\Bstrut & 204 & 61 & 397 & 59& 253 & 75 & 184 & 63\\
\textbf{Ours}    & 107 & 88 & 240 & 40& 285 & 59 & 130 & 61  \\
\hline
\end{tabular}\\  
\Tstrut\Bstrut
\caption{Mean and Standard Deviation evaluation of inliers and outliers between various baselines and ours. Our method quantitatively reduces the number of outliers in the majority of the baselines. As we reduce the search space, the inlier candidates also get reduced. We show that this inlier reduction works in our favor in section~\ref{sec:evaluation}.}
\label{table2}
\vspace*{-3em}
\end{center}
\end{table*}
\subsection{Visual Odometry Estimation}\label{sec:4b}
RANSAC being a non-deterministic technique of estimating essential matrix \cite{zhang1998determining} which helps in recovering a robot's relative pose, we show the efficacy of the inliers and outliers obtained using our method by performing trajectory estimation using our feature space. In our method, we reduce the number of candidate features by restricting the image to regions of interest. In a 2d space $W \times H$, our feature space corresponds to $M \times N$ $\in$ $W \times H$ with a consistent reduction parameter of $1/n$ over certain sequences of images. This search space reduces the overall distribution of features by a margin factor $\lambda$.
\begin{table*}
\centering
  \begin{tabular}{|l|l|l|l|l|l|l|l|l|l|l|}
    \hline
    \multirow{2}{*}{Detectors} &
      \multicolumn{4}{c}{KITTI \cite{geiger2013vision}} &
      \multicolumn{4}{c}{EuRoc \cite{burri2016euroc}}   \\ 
    & t-error & R-error & Pose Error & Rep Error& Time & t-error & R-error & Pose Error &Rep Error& Time\\
    \hline

    ORB \cite{mur2015orb} \Tstrut\Bstrut & 3.37 & 0.02 & 1.5 &82 & 49.68 &3.22 & 0.04& 2.5 &37&45.54 \\
    \textbf{Ours}  & 3.11 & 0.01 & \textbf{1.5}& \textbf{81}& \textbf{43.52} & 3.19 & 0.04& \textbf{2.4} &\textbf{36}& \textbf{43.44}   \\
    \hline
    SIFT \cite{alhwarin2010vf} \Tstrut\Bstrut & 3.30 & 0.02 & 1.4 & 100 &44.90 & 3.86 & 0.04 & 2.4 &55&38.46 \\
    \textbf{Ours}  & 3.16 & 0.02 & \textbf{1.1} & \textbf{98}&\textbf{40.11} & 3.79 & 0.05 & \textbf{2.3} & \textbf{52}&\textbf{36.44}  \\
    \hline
    FAST \cite{chiu2013fast} \Tstrut\Bstrut & 3.05 & 0.02 & 1.5 & 120 &28.82 & 2.97 & 0.04 & 2.4 &88& 36.38 \\
    \textbf{Ours}  & 2.87 & 0.02 & \textbf{1.4} & 128&\textbf{28.22} & 2.58 &0.04& \textbf{2.4}  &95&\textbf{35.94} \\
    \hline
    KAZE \cite{alcantarilla2012kaze} \Tstrut\Bstrut & 2.79 & 0.02 & 1.4 & 143&578.5 &3.14& 0.04 &2.5&134&448.6 \\
    \textbf{Ours}  & 2.59 & 0.02 & \textbf{0.9} & \textbf{139} &\textbf{558.2} & 3.06 & 0.04 & \textbf{2.4} & 134 &\textbf{443.1}  \\
    \hline
    AKAZE \cite{sharma2020image} \Tstrut\Bstrut & 2.46 & 0.01 & 2.1 & 92&38.59 & 2.68 & 0.04 & 2.3&83&37.17 \\
    \textbf{Ours}  & 2.29 & 0.02 & \textbf{2.0} & \textbf{90}&\textbf{36.40} & 2.46 & 0.04 & \textbf{2.0} & \textbf{87}&\textbf{35.33} \\
    \hline
    BRISK \cite{leutenegger2011brisk} \Tstrut\Bstrut & 2.78 & 0.03 & 1.2 &102 &48.32 & 2.89 & 0.03 & 2.5&73&45.54 \\
    \textbf{Ours}  & 2.66 & 0.03 & \textbf{0.9} & \textbf{99}&\textbf{45.68} & 2.51 &0.04& \textbf{2.2}&\textbf{70}&\textbf{44.22}   \\
    \hline
    SuperPoint \cite{detone2018superpoint} \Tstrut\Bstrut & 2.73 & 0.03 & 1.7 & 51&278.7 & 2.24 &  0.03 & 2.2 &46&260.58\\
    \textbf{Ours}  & 2.61 & 0.03 & \textbf{1.6} & \textbf{50}&\textbf{262.3} & 2.01 & 0.03 & \textbf{2.1} &\textbf{44} &\textbf{263.67}  \\
    \hline
  
  \end{tabular}\\
    \Tstrut\Bstrut
    \vspace*{-1em}
    \caption{Trajectory error evaluation representing translation, rotational and pose error (m) followed by the Average Reprojection Error (Pixels) and Average Execution Time (s) using baseline features overlayed on original image spaces of \kitty~(376x1241) and \euroc~(480x752) vs using baseline features overlayed on our attentive space.}
\label{table3}
\end{table*}

In our search space, given a pair of corresponding points $\lambda p$  and $\lambda p'$ in normalized camera coordinates (where $p$ and $p'$ correspond to the feature points obtained using any baseline shown in table~\ref{table2} over the complete image area $W \times H$, the equation for the Essential matrix will be $\lambda(p'^TEp)$ = 0. where $E$ is the essential matrix from \cite{zhang1998determining}. Before estimating the current pose, we refine the poorly fitted point pairs using RANSAC. We observe that our search space $(M \times N)$ results in a reduced number of outliers in most of the baselines as shown in table~\ref{table2}. Using the current pose as an anchor node, we estimate all the successive poses using $E = [t]_x * R$ and decompose all the R(rotations) and t(translations) using SVD (Singular Value Decomposition) $E = U * W * V^T$ , where U, V are unitary matrices

W is a diagonal matrix. To calculate the Average Trajectory Error between all the poses obtained using our method and the ground truth, we calculate RMSE (Root Mean Square) between the estimated poses and ground truth poses. We evaluate ATE using image sequences from test data and the results are shown in table~\ref{table3}. We observe that our approach focuses on the reduction of image space without compromising the trajectory estimation accuracy.

\label{sec:result}
\subsection{Reprojection Error}\label{sec:4c}
We support the inlier selection overlayed on our attention space by calculating the reprojection errors using homography metric between all incremental image sequences. Table~\ref{table3} represents the pixels accounting for reprojection errors after triangulation of corresponding feature points.  




To record the change in positional uncertainties, we create covariance matrixes as mentioned in \cite{muhle2023learning}. The baseline-based feature detector and descriptor extractor are created. Key points and descriptors are computed for img1 and img2 using the ORB detector and extractor. Baseline descriptors overlayed into the image space inferred from L-DYNO are matched between img1 and img2 using the BFMatcher, and matches are sorted by distance. We calculate a homography matrix H using RANSAC. This matrix estimates the geometric transformation between keypoints in sequential image pairs. Next, we calculate the reprojection error for each key point in image n after applying the estimated homography to project them into the image n+1 space. This error quantifies the dissimilarity between matched points.





\subsection{Execution Time}\label{sec:4d}
Another axis in which our approach yields positive results is the execution time of the VO pipeline. Intuitively, reduced feature candidates from our method should result in reduced time taken by the $E$ matrix estimation. We take account of this evaluation by comparing the execution time spent by the VO pipeline from PYSLAMv2 \cite{fredaPySLAMV2} module to perform an end-to-end pose estimation using our search space area and compare it to other methods in table~\ref{table3}. We experienced a noticeable drop of an average of 8.51 seconds in the time taken by our method to execute the complete pipeline. This validation supports our initial motive to refine the features and select the features that are \textit{most important}. 

\subsection{Real World Robot Experiment}\label{sec:f1tenth}
\begin{figure}
    \includegraphics[width=8.5cm, height=6cm]{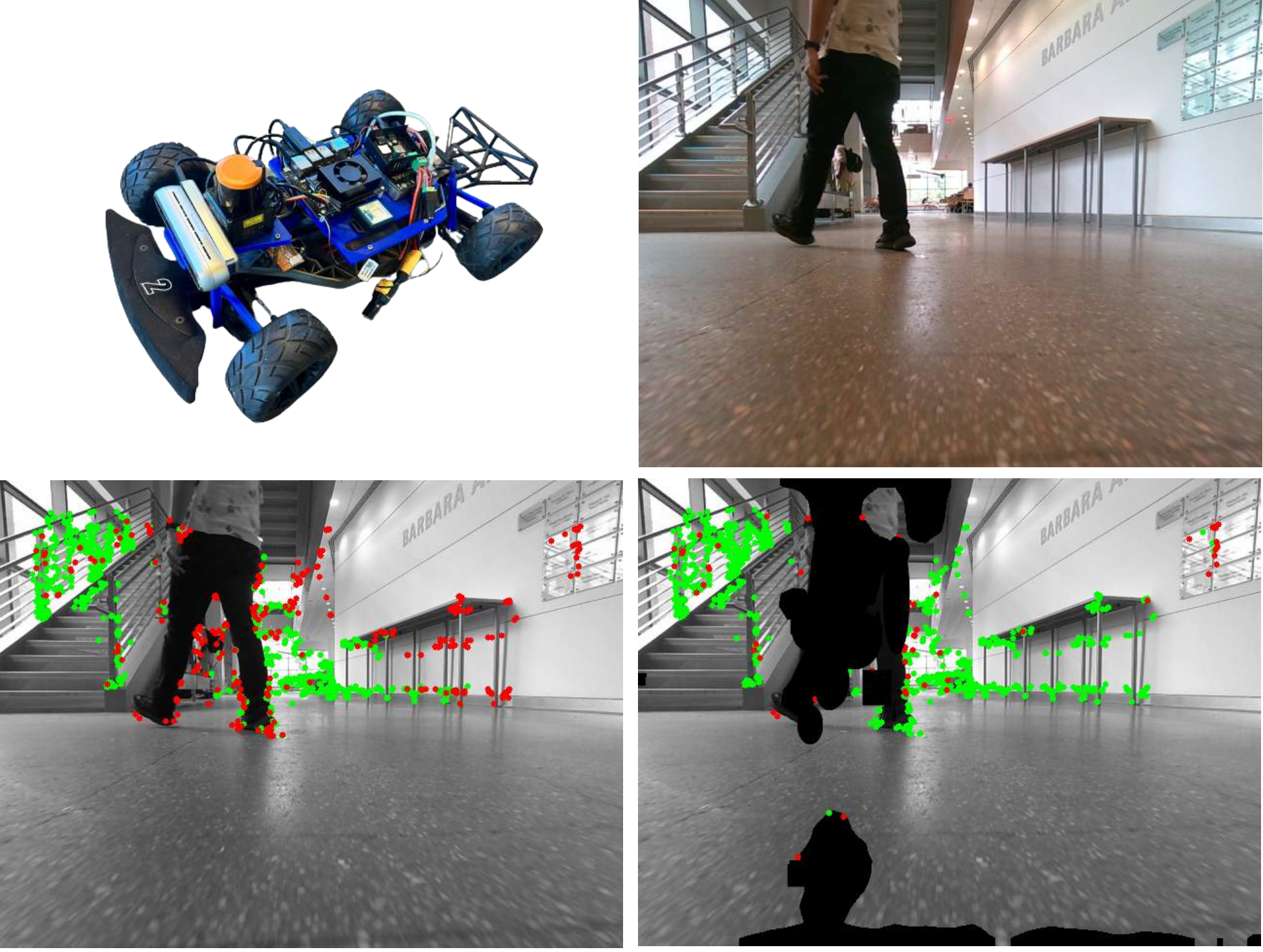}
    \caption{Experiment Setup(Top Left), Raw Image(Top Right) with dynamic objects inducing outliers using an orthogonal range of movements. Inliers vs Outliers distribution on raw image space vs L-DYNO(Bottom).}
    \label{fig:f1tenth}
\end{figure}

We demonstrate the performance of L-DYNO using a custom odometry dataset. Figure~\ref{fig:f1tenth} represents the overall setup used in recording RGB (Realsense D455), IMU (Realsense D455), and ground truth data (Realsense t265). Table~\ref{table4} represents the quantitative results based on the evaluation methods mentioned in Section~\ref{sec:4b} \ref{sec:4c}, and \ref{sec:4d}. We induce inconsistency in the scene by having a person walk across the robot as shown in Figure~\ref{fig:f1tenth}. These sudden movements  produce more outliers in a standard case. The training data contains over 1006 images of size 640x480 sequences with corresponding IMU and t265 base ground truth poses. L-DYNO reduces the image space by \textbf{20\%} for all the 300 completely unseen test images.  
Additionally, in-order to validate the performance of L-DYNO under the high influence of noise in IMU, we avoid using the window-based $(W)$ approach to form a trajectory using IMU. Table~\ref{table4} shows the improvements in the pose, reprojection errors, and time when we compare L-DYNO image space using baseline detectors. 

\begin{table}[]
\begin{tabular}{|l|c|c|c|}
\hline
Detectors &
  \multicolumn{1}{l|}{Pose Error(mm)} &
  \multicolumn{1}{l|}{Rep Error(Pixel)} &
  \multicolumn{1}{l|}{Time(Sec)} \\ \hline
\begin{tabular}[c]{@{}l@{}}SIFT\\ Ours\end{tabular} &
  \begin{tabular}[c]{@{}c@{}}49\\ \textbf{22}\end{tabular} &
  \begin{tabular}[c]{@{}c@{}}85\\ \textbf{63}\end{tabular} &
  \begin{tabular}[c]{@{}c@{}}20\\ \textbf{18}\end{tabular} \\ \hline
\begin{tabular}[c]{@{}l@{}}ORB\\ Ours\end{tabular} &
  \begin{tabular}[c]{@{}c@{}}23\\ 57\end{tabular} &
  \begin{tabular}[c]{@{}c@{}}58\\ 61\end{tabular} &
  \begin{tabular}[c]{@{}c@{}}10\\ 10\end{tabular} \\ \hline
\begin{tabular}[c]{@{}l@{}}Fast\\ Ours\end{tabular} &
  \begin{tabular}[c]{@{}c@{}}63\\ \textbf{23}\end{tabular} &
  \begin{tabular}[c]{@{}c@{}}61\\ \textbf{45}\end{tabular} &
  \begin{tabular}[c]{@{}c@{}}11\\ \textbf{9}\end{tabular} \\ \hline
\begin{tabular}[c]{@{}l@{}}KAZE\\ Ours\end{tabular} &
  \begin{tabular}[c]{@{}c@{}}62\\ \textbf{30}\end{tabular} &
  \begin{tabular}[c]{@{}c@{}}62\\ \textbf{46}\end{tabular} &
  \begin{tabular}[c]{@{}c@{}}46\\ \textbf{42}\end{tabular} \\ \hline
\begin{tabular}[c]{@{}l@{}}AKAZE\\ Ours\end{tabular} &
  \begin{tabular}[c]{@{}c@{}}24\\ \textbf{5}\end{tabular} &
  \begin{tabular}[c]{@{}c@{}}39\\ \textbf{26}\end{tabular} &
  \begin{tabular}[c]{@{}c@{}}17\\ \textbf{15}\end{tabular} \\ \hline
\begin{tabular}[c]{@{}l@{}}BRISK\\ Ours\end{tabular} &
  \begin{tabular}[c]{@{}c@{}}41\\ \textbf{38}\end{tabular} &
  \begin{tabular}[c]{@{}c@{}}77\\ \textbf{75}\end{tabular} &
  \begin{tabular}[c]{@{}c@{}}14\\ \textbf{13}\end{tabular} \\ \hline
\begin{tabular}[c]{@{}l@{}}SUPERPOINT\\ Ours\end{tabular} &
  \begin{tabular}[c]{@{}c@{}}22\\ \textbf{21}\end{tabular} &
  \begin{tabular}[c]{@{}c@{}}34\\ \textbf{31}\end{tabular} &
  \begin{tabular}[c]{@{}c@{}}39\\ \textbf{36}\end{tabular} \\ \hline
\end{tabular}
\caption{Evaluations on the real robot data recorded. We notice a big noticeable reduction in Pose and reprojection errors and optimized inference time.}
\label{table4}
\end{table}

\section{Conclusion}\label{sec:conclusion}
We present a representation learning-based approach that learns the region of interest for selecting consistent features in order to execute accurate camera-based perception tasks like localization, mapping, and tracking. We demonstrate how to leverage the information from robot's motion and use it to supervise the training network. By doing this, we investigate the possibilities of bridging the computer vision and robotics domains together and utilize the interchangeable signals from each domain using representation learning. We demonstrate a method to accomplish consistent feature selection with reduced outlier candidates. Our method achieves improved trajectory estimations, reduced reprojection errors, and shows improved results in execution time even after reducing the image space by \textbf{49\%}. Our method shows substantial improvements over all the evaluations performed using both real-world datasets like KITTI \cite{geiger2013vision} and EuRoC \cite{burri2016euroc} as well as dataset recorded from a real robot.  


\bibliographystyle{IEEEtran}
\bibliography{references}

\

\end{document}

%% file: meta/usercmds.tex
\newcommand{\kitty}[1]{KITTI}
\newcommand{\euroc}[1]{EuRoC}
\newcommand{\imu}[1]{IMU}
\newcommand{\teaser}[1]{TEASER++}

%% file: root.bbl
\begin{thebibliography}{10}
\providecommand{\url}[1]{#1}
\csname url@samestyle\endcsname
\providecommand{\newblock}{\relax}
\providecommand{\bibinfo}[2]{#2}
\providecommand{\BIBentrySTDinterwordspacing}{\spaceskip=0pt\relax}
\providecommand{\BIBentryALTinterwordstretchfactor}{4}
\providecommand{\BIBentryALTinterwordspacing}{\spaceskip=\fontdimen2\font plus
\BIBentryALTinterwordstretchfactor\fontdimen3\font minus
  \fontdimen4\font\relax}
\providecommand{\BIBforeignlanguage}[2]{{%
\expandafter\ifx\csname l@#1\endcsname\relax
\typeout{** WARNING: IEEEtran.bst: No hyphenation pattern has been}%
\typeout{** loaded for the language `#1'. Using the pattern for}%
\typeout{** the default language instead.}%
\else
\language=\csname l@#1\endcsname
\fi
#2}}
\providecommand{\BIBdecl}{\relax}
\BIBdecl

\bibitem{raguram2008comparative}
R.~Raguram, J.-M. Frahm, and M.~Pollefeys, ``A comparative analysis of ransac
  techniques leading to adaptive real-time random sample consensus,'' in
  \emph{Computer Vision--ECCV 2008: 10th European Conference on Computer
  Vision, Marseille, France, October 12-18, 2008, Proceedings, Part II
  10}.\hskip 1em plus 0.5em minus 0.4em\relax Springer, 2008, pp. 500--513.

\bibitem{yang2020teaser}
H.~Yang, J.~Shi, and L.~Carlone, ``Teaser: Fast and certifiable point cloud
  registration,'' \emph{IEEE Transactions on Robotics}, vol.~37, no.~2, pp.
  314--333, 2020.

\bibitem{jiang2021cotr}
W.~Jiang, E.~Trulls, J.~Hosang, A.~Tagliasacchi, and K.~M. Yi, ``Cotr:
  Correspondence transformer for matching across images,'' in \emph{Proceedings
  of the IEEE/CVF International Conference on Computer Vision}, 2021, pp.
  6207--6217.

\bibitem{mao20223dg}
R.~Mao, C.~Bai, Y.~An, F.~Zhu, and C.~Lu, ``3dg-stfm: 3d geometric guided
  student-teacher feature matching,'' in \emph{Computer Vision--ECCV 2022: 17th
  European Conference, Tel Aviv, Israel, October 23--27, 2022, Proceedings,
  Part XXVIII}.\hskip 1em plus 0.5em minus 0.4em\relax Springer, 2022, pp.
  125--142.

\bibitem{wei2021unsupervised}
P.~Wei, G.~Hua, W.~Huang, F.~Meng, and H.~Liu, ``Unsupervised monocular
  visual-inertial odometry network,'' in \emph{Proceedings of the Twenty-Ninth
  International Conference on International Joint Conferences on Artificial
  Intelligence}, 2021, pp. 2347--2354.

\bibitem{geiger2013vision}
A.~Geiger, P.~Lenz, C.~Stiller, and R.~Urtasun, ``Vision meets robotics: The
  kitti dataset,'' \emph{The International Journal of Robotics Research},
  vol.~32, no.~11, pp. 1231--1237, 2013.

\bibitem{burri2016euroc}
M.~Burri, J.~Nikolic, P.~Gohl, T.~Schneider, J.~Rehder, S.~Omari, M.~W.
  Achtelik, and R.~Siegwart, ``The euroc micro aerial vehicle datasets,''
  \emph{The International Journal of Robotics Research}, vol.~35, no.~10, pp.
  1157--1163, 2016.

\bibitem{o2020f1tenth}
M.~O'Kelly, H.~Zheng, D.~Karthik, and R.~Mangharam, ``F1tenth: An open-source
  evaluation environment for continuous control and reinforcement learning,''
  \emph{Proceedings of Machine Learning Research}, vol. 123, 2020.

\bibitem{derpanis2004harris}
K.~G. Derpanis, ``The harris corner detector,'' \emph{York University}, vol.~2,
  pp. 1--2, 2004.

\bibitem{harris1988combined}
C.~Harris, M.~Stephens \emph{et~al.}, ``A combined corner and edge detector,''
  in \emph{Alvey vision conference}, vol.~15, no.~50.\hskip 1em plus 0.5em
  minus 0.4em\relax Citeseer, 1988, pp. 10--5244.

\bibitem{carron1994color}
T.~Carron and P.~Lambert, ``Color edge detector using jointly hue, saturation
  and intensity,'' in \emph{Proceedings of 1st International Conference on
  Image Processing}, vol.~3.\hskip 1em plus 0.5em minus 0.4em\relax IEEE, 1994,
  pp. 977--981.

\bibitem{baker2004lucas}
S.~Baker and I.~Matthews, ``Lucas-kanade 20 years on: A unifying framework,''
  \emph{International journal of computer vision}, vol.~56, pp. 221--255, 2004.

\bibitem{garcia2010k}
V.~Garcia, E.~Debreuve, F.~Nielsen, and M.~Barlaud, ``K-nearest neighbor
  search: Fast gpu-based implementations and application to high-dimensional
  feature matching,'' in \emph{2010 IEEE International Conference on Image
  Processing}.\hskip 1em plus 0.5em minus 0.4em\relax IEEE, 2010, pp.
  3757--3760.

\bibitem{vijayan2019flann}
V.~Vijayan and P.~Kp, ``Flann based matching with sift descriptors for drowsy
  features extraction,'' in \emph{2019 Fifth International Conference on Image
  Information Processing (ICIIP)}.\hskip 1em plus 0.5em minus 0.4em\relax IEEE,
  2019, pp. 600--605.

\bibitem{younes2017keyframe}
G.~Younes, D.~Asmar, E.~Shammas, and J.~Zelek, ``Keyframe-based monocular slam:
  design, survey, and future directions,'' \emph{Robotics and Autonomous
  Systems}, vol.~98, pp. 67--88, 2017.

\bibitem{mur2015orb}
R.~Mur-Artal, J.~M.~M. Montiel, and J.~D. Tardos, ``Orb-slam: a versatile and
  accurate monocular slam system,'' \emph{IEEE transactions on robotics},
  vol.~31, no.~5, pp. 1147--1163, 2015.

\bibitem{engel2014lsd}
J.~Engel, T.~Sch{\"o}ps, and D.~Cremers, ``Lsd-slam: Large-scale direct
  monocular slam,'' in \emph{Computer Vision--ECCV 2014: 13th European
  Conference, Zurich, Switzerland, September 6-12, 2014, Proceedings, Part II
  13}.\hskip 1em plus 0.5em minus 0.4em\relax Springer, 2014, pp. 834--849.

\bibitem{forster2014svo}
C.~Forster, M.~Pizzoli, and D.~Scaramuzza, ``Svo: Fast semi-direct monocular
  visual odometry,'' in \emph{2014 IEEE international conference on robotics
  and automation (ICRA)}.\hskip 1em plus 0.5em minus 0.4em\relax IEEE, 2014,
  pp. 15--22.

\bibitem{engel2017direct}
J.~Engel, V.~Koltun, and D.~Cremers, ``Direct sparse odometry,'' \emph{IEEE
  transactions on pattern analysis and machine intelligence}, vol.~40, no.~3,
  pp. 611--625, 2017.

\bibitem{mohanty2016deepvo}
V.~Mohanty, S.~Agrawal, S.~Datta, A.~Ghosh, V.~D. Sharma, and D.~Chakravarty,
  ``Deepvo: A deep learning approach for monocular visual odometry,''
  \emph{arXiv preprint arXiv:1611.06069}, 2016.

\bibitem{wang2021tartanvo}
W.~Wang, Y.~Hu, and S.~Scherer, ``Tartanvo: A generalizable learning-based
  vo,'' in \emph{Conference on Robot Learning}.\hskip 1em plus 0.5em minus
  0.4em\relax PMLR, 2021, pp. 1761--1772.

\bibitem{gao2022unsupervised}
R.~Gao, X.~Xiao, W.~Xing, C.~Li, and L.~Liu, ``Unsupervised learning of
  monocular depth and ego-motion in outdoor/indoor environments,'' \emph{IEEE
  Internet of Things Journal}, vol.~9, no.~17, pp. 16\,247--16\,258, 2022.

\bibitem{vijayanarasimhan2017sfm}
S.~Vijayanarasimhan, S.~Ricco, C.~Schmid, R.~Sukthankar, and K.~Fragkiadaki,
  ``Sfm-net: Learning of structure and motion from video,'' \emph{arXiv
  preprint arXiv:1704.07804}, 2017.

\bibitem{li2018undeepvo}
R.~Li, S.~Wang, Z.~Long, and D.~Gu, ``Undeepvo: Monocular visual odometry
  through unsupervised deep learning,'' in \emph{2018 IEEE international
  conference on robotics and automation (ICRA)}.\hskip 1em plus 0.5em minus
  0.4em\relax IEEE, 2018, pp. 7286--7291.

\bibitem{yin2018geonet}
Z.~Yin and J.~Shi, ``Geonet: Unsupervised learning of dense depth, optical flow
  and camera pose,'' in \emph{Proceedings of the IEEE conference on computer
  vision and pattern recognition}, 2018, pp. 1983--1992.

\bibitem{wang2019unos}
Y.~Wang, P.~Wang, Z.~Yang, C.~Luo, Y.~Yang, and W.~Xu, ``Unos: Unified
  unsupervised optical-flow and stereo-depth estimation by watching videos,''
  in \emph{Proceedings of the IEEE/CVF conference on computer vision and
  pattern recognition}, 2019, pp. 8071--8081.

\bibitem{chiu2013fast}
L.-C. Chiu, T.-S. Chang, J.-Y. Chen, and N.~Y.-C. Chang, ``Fast sift design for
  real-time visual feature extraction,'' \emph{IEEE Transactions on Image
  Processing}, vol.~22, no.~8, pp. 3158--3167, 2013.

\bibitem{detone2018superpoint}
D.~DeTone, T.~Malisiewicz, and A.~Rabinovich, ``Superpoint: Self-supervised
  interest point detection and description,'' in \emph{Proceedings of the IEEE
  conference on computer vision and pattern recognition workshops}, 2018, pp.
  224--236.

\bibitem{sun2018pwc}
D.~Sun, X.~Yang, M.-Y. Liu, and J.~Kautz, ``Pwc-net: Cnns for optical flow
  using pyramid, warping, and cost volume,'' in \emph{Proceedings of the IEEE
  conference on computer vision and pattern recognition}, 2018, pp. 8934--8943.

\bibitem{sarlin2020superglue}
P.-E. Sarlin, D.~DeTone, T.~Malisiewicz, and A.~Rabinovich, ``Superglue:
  Learning feature matching with graph neural networks,'' in \emph{Proceedings
  of the IEEE/CVF conference on computer vision and pattern recognition}, 2020,
  pp. 4938--4947.

\bibitem{sun2021loftr}
J.~Sun, Z.~Shen, Y.~Wang, H.~Bao, and X.~Zhou, ``Loftr: Detector-free local
  feature matching with transformers,'' in \emph{Proceedings of the IEEE/CVF
  conference on computer vision and pattern recognition}, 2021, pp. 8922--8931.

\bibitem{xie2021cotr}
Y.~Xie, J.~Zhang, C.~Shen, and Y.~Xia, ``Cotr: Efficiently bridging cnn and
  transformer for 3d medical image segmentation,'' in \emph{Medical Image
  Computing and Computer Assisted Intervention--MICCAI 2021: 24th International
  Conference, Strasbourg, France, September 27--October 1, 2021, Proceedings,
  Part III 24}.\hskip 1em plus 0.5em minus 0.4em\relax Springer, 2021, pp.
  171--180.

\bibitem{roessle2022end2end}
B.~Roessle and M.~Nie{\ss}ner, ``End2end multi-view feature matching using
  differentiable pose optimization,'' \emph{arXiv preprint arXiv:2205.01694},
  2022.

\bibitem{alhwarin2010vf}
F.~Alhwarin, D.~Risti{\'c}-Durrant, and A.~Gr{\"a}ser, ``Vf-sift: very fast
  sift feature matching,'' in \emph{Pattern Recognition: 32nd DAGM Symposium,
  Darmstadt, Germany, September 22-24, 2010. Proceedings 32}.\hskip 1em plus
  0.5em minus 0.4em\relax Springer, 2010, pp. 222--231.

\bibitem{leutenegger2011brisk}
S.~Leutenegger, M.~Chli, and R.~Y. Siegwart, ``Brisk: Binary robust invariant
  scalable keypoints,'' in \emph{2011 International conference on computer
  vision}.\hskip 1em plus 0.5em minus 0.4em\relax Ieee, 2011, pp. 2548--2555.

\bibitem{sharma2020image}
S.~K. Sharma and K.~Jain, ``Image stitching using akaze features,''
  \emph{Journal of the Indian Society of Remote Sensing}, vol.~48, pp.
  1389--1401, 2020.

\bibitem{alcantarilla2012kaze}
P.~F. Alcantarilla, A.~Bartoli, and A.~J. Davison, ``Kaze features,'' in
  \emph{Computer Vision--ECCV 2012: 12th European Conference on Computer
  Vision, Florence, Italy, October 7-13, 2012, Proceedings, Part VI 12}.\hskip
  1em plus 0.5em minus 0.4em\relax Springer, 2012, pp. 214--227.

\bibitem{muja2012fast}
M.~Muja and D.~G. Lowe, ``Fast matching of binary features,'' in \emph{2012
  Ninth conference on computer and robot vision}.\hskip 1em plus 0.5em minus
  0.4em\relax IEEE, 2012, pp. 404--410.

\bibitem{zhang1998determining}
Z.~Zhang, ``Determining the epipolar geometry and its uncertainty: A review,''
  \emph{International journal of computer vision}, vol.~27, pp. 161--195, 1998.

\bibitem{muhle2023learning}
D.~Muhle, L.~Koestler, K.~M. Jatavallabhula, and D.~Cremers, ``Learning
  correspondence uncertainty via differentiable nonlinear least squares,'' in
  \emph{Proceedings of the IEEE/CVF Conference on Computer Vision and Pattern
  Recognition}, 2023, pp. 13\,102--13\,112.

\bibitem{fredaPySLAMV2}
\BIBentryALTinterwordspacing
L.~Freda, ``{pySLAM} {V2}.'' [Online]. Available:
  \url{https://github.com/luigifreda/pyslam}
\BIBentrySTDinterwordspacing

\end{thebibliography}
